\documentclass[final,3p,times,twocolumn]{elsarticle}
\biboptions{sort}          % 不要排序就去掉这行

\usepackage{amssymb}
\usepackage{amsmath}
\usepackage{graphicx}
\usepackage{subcaption}
\usepackage{booktabs}
\usepackage{multirow}
\usepackage{float}
\usepackage{makecell}
\usepackage{color}
\usepackage{xspace}
\usepackage{bm}
\usepackage{threeparttable}
\usepackage{algorithm}
\usepackage{algpseudocode}

\usepackage{xcolor}
\usepackage{hyperref}
\hypersetup{
  colorlinks=true,
  linkcolor=blue,
  citecolor=blue,
  urlcolor=blue
}
\journal{Computerized Medical Imaging and Graphics}

\begin{document}

\begin{frontmatter}



\title{Real-Time Glottis Detection Framework via Spatial-decoupled Feature Learning for Nasal Transnasal Intubation}


\author[1]{Jinyu Liu}
\author[1]{Gaoyang Zhang}
\author[2]{Yang Zhou}
\author[3]{Ruoyi Hao}
\author[1,4,5]{Yang Zhang\textsuperscript{*}}
\author[3]{Hongliang Ren\textsuperscript{**}}

\cortext[cor1]{Corresponding author: Yang Zhang (email: yzhangcst@hbut.edu.cn)}
\cortext[cor2]{Corresponding author: Hongliang Ren (email: hlren@ee.cuhk.edu.hk)}

\affiliation[1]{organization={Hubei Key Laboratory of Modern Manufacturing Quality Engineering, Hubei University of Technology},
    addressline={Nanhu Avenue 28}, 
    city={Wuhan},
    postcode={430068}, 
    state={Hubei},
    country={China}
}

\affiliation[2]{organization={School of Mechanical Science and Engineering, Huazhong University of Science and Technology},
    addressline={Luoyu Road 1037}, 
    city={Wuhan},
    postcode={430074}, 
    state={Hubei},
    country={China}
}

\affiliation[3]{organization={Department of Electronic Engineering, The Chinese University of Hong Kong},
    addressline={Shatin}, 
    city={Hong Kong},
    postcode={999077}, 
    country={China}
}

\affiliation[4]{organization={Key Laboratory of Symbolic Computation and Knowledge Engineering, Ministry of Education},
    addressline={Jilin University}, 
    city={Changchun},
    postcode={130012}, 
    state={Jilin},
    country={China}
}

\affiliation[5]{organization={National Key Laboratory for Novel Software Technology, Nanjing University},
    addressline={Hankou Road 22}, 
    city={Nanjing},
    postcode={210023}, 
    state={Jiangsu},
    country={China}
}

%% Abstract
\begin{abstract}
Nasotracheal intubation (NTI) is a vital procedure in emergency airway management, where rapid and accurate glottis detection is essential to ensure patient safety. However, existing machine assisted visual detection systems often rely on high performance computational resources and suffer from significant inference delays, which limits their applicability in time critical and resource constrained scenarios. To overcome these limitations, we propose Mobile GlottisNet, a lightweight and efficient glottis detection framework designed for real time inference on embedded and edge devices. The model incorporates structural awareness and spatial alignment mechanisms, enabling robust glottis localization under complex anatomical and visual conditions. We implement a hierarchical dynamic thresholding strategy to enhance sample assignment, and introduce an adaptive feature decoupling module based on deformable convolution to support dynamic spatial reconstruction. A cross layer dynamic weighting scheme further facilitates the fusion of semantic and detail features across multiple scales. Experimental results demonstrate that the model, with a size of only 5MB on both our PID dataset and Clinical datasets, achieves inference speeds of over 62 FPS on devices and 33 FPS on edge platforms, showing great potential in the application of emergency NTI.
\end{abstract}

%% Keywords

\end{frontmatter}



\section{Introduction}
\label{sec:introduction}
Airway management is one of the most critical interventions in emergency medicine, where precise identification and access to glottic structures directly determine patient survival outcomes~\cite{vaz2017airway}. In particular, nasotracheal intubation (NTI) poses greater technical challenges than orotracheal intubation, as the endotracheal tube must be advanced through a narrower and longer nasal pathway with limited visualization~\cite{deng2024assisted}. Adverse conditions such as poor illumination, anatomical variability, secretion interference, and patient movement further complicate the process, underscoring the critical importance of accurate real-time glottis localization. Traditional intubation procedures rely heavily on the operator’s expertise and tactile feedback for blind insertion, posing significant risks in emergency situations. Studies~\cite{ryoo2024utilization} have shown that even among experienced practitioners, the first-attempt success rate for emergency intubation can be as low as 70--80\%, with failure rates rising sharply under adverse conditions. These limitations have motivated the development of computer vision-assisted and robotic-assisted intubation systems to provide real-time visual guidance, thereby improving glottic exposure and reducing complications.

Recent advances in robotic-assisted intubation systems have demonstrated the potential for autonomous glottis detection and tube guidance~\cite{chen2025emerging}. However, these platforms typically rely on computationally intensive deep learning algorithms, high-resolution imaging devices, and stable power supplies, leading to bulky configurations and high costs that hinder their use in prehospital emergencies and resource-limited settings. Similarly, glottis-related computer vision tasks have often been formulated as semantic segmentation problems, which provide pixel-level precision but require high-resolution feature maps and substantial memory, making them impractical for embedded emergency devices. In contrast, detection frameworks based on bounding box localization offer a more pragmatic solution, delivering sufficient spatial precision for trajectory planning and robotic NTI guidance while maintaining computational efficiency suitable for edge deployment. Nevertheless, current detectors still face fundamental challenges in real-world clinical settings, including motion blur from patient movement, secretion or blood occlusion, fogging of optical surfaces, and extreme lighting variations. Moreover, their complex architectures often result in inference delays exceeding 200 ms, conflicting with the stringent time requirements of emergency airway management, where the golden window for successful intervention is only 3–5 minutes~\cite{liaqat2025difficult}. These limitations highlight the urgent need for lightweight and efficient algorithms capable of delivering clinical-grade accuracy under the constraints of mobile emergency environments.

\begin{figure*}[!t]
  \centering
  \setlength{\tabcolsep}{1pt}
  \renewcommand{\arraystretch}{1}

  \includegraphics[width=5in]{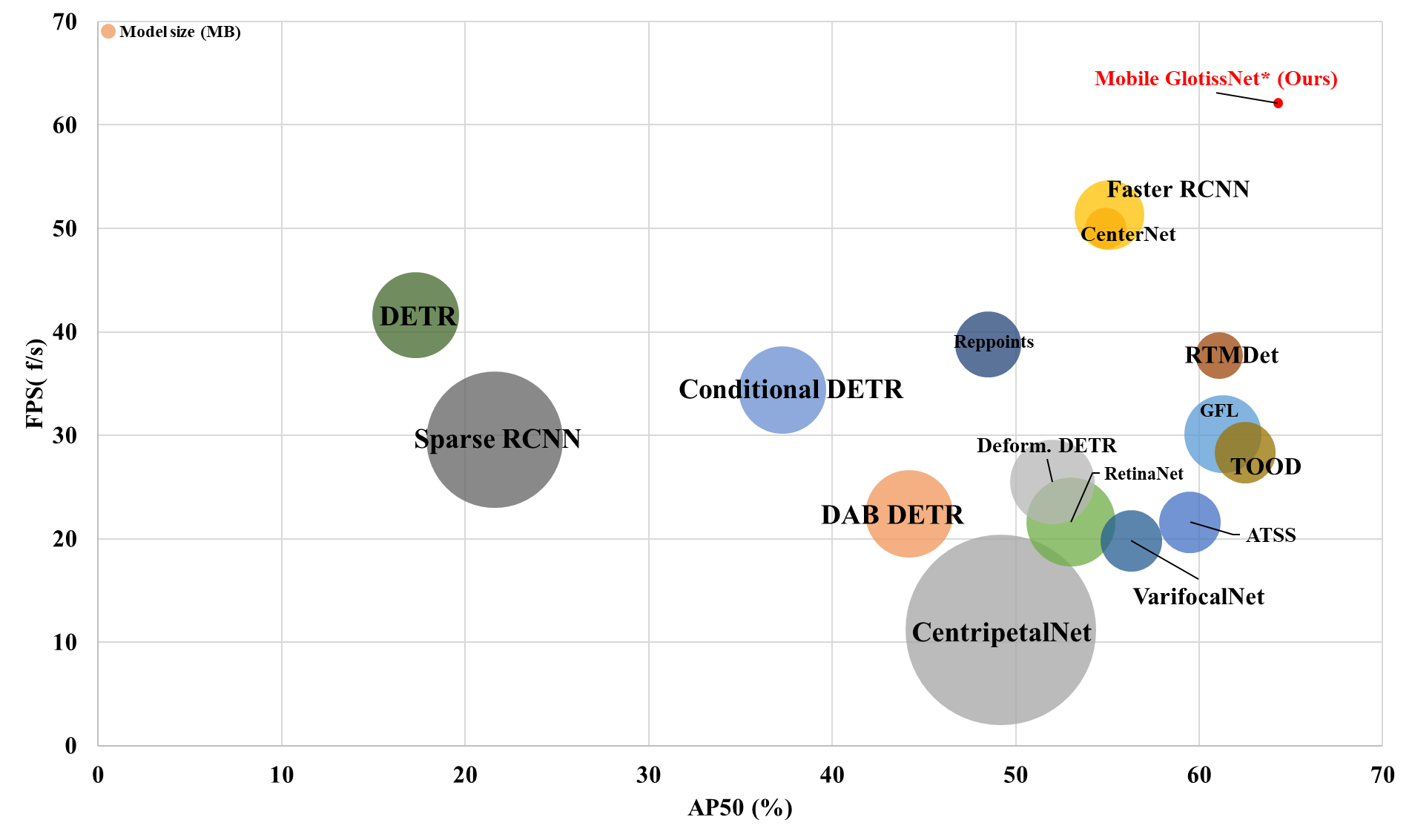}

  \caption{\scriptsize The comparative visualization with state-of-the-art methods on the PID dataset is illustrated, where the vertical axis represents the inference speed (FPS) measured on terminal devices. The bubble chart shows that our method achieves a favorable balance between AP50 and FPS while maintaining a compact model size. The area of each bubble corresponds to the model size.}
  \label{PID-FPS}
\end{figure*}

To address the limitations of existing technologies, we propose Mobile GlottisNet, a lightweight glottis detection framework for NTI, optimized for both performance and hardware adaptability. By integrating dynamic thresholding and hierarchical selection, the allocation of samples for classification and regression tasks is optimized. An adaptive feature decoupling module enhances model robustness under complex conditions such as blood, fog, and low light by learning distinct spatial regions in the glottal feature map. An adaptive weighting mechanism further adjusts supervision strength across different features. The innovative use of a lightweight backbone significantly reduces computational complexity while preserving high detection accuracy, enabling efficient inference on low-power embedded platforms. As illustrated in Fig.~\ref{PID-FPS}, Mobile GlottisNet achieves a compact model size of only 5 MB while delivering real-time inference speeds exceeding 62 FPS on terminal devices and 33 FPS on edge platforms, striking a favorable trade-off between accuracy and efficiency compared with state-of-the-art methods. This computational advantage is particularly critical for resource-constrained deployment, including bedside endoscopic equipment, cost-effective training simulators for medical education, and portable robotic-assisted airway devices for emergency interventions such as wilderness rescue. 

The main contributions of this work are as follows.
\begin{enumerate}
    \item We propose a lightweight glottis detection framework with structural awareness and spatial alignment tailored for nasotracheal intubation, enabling efficient and accurate localization of the glottis in resource-constrained scenarios.
    
    \item A hierarchical dynamic thresholding strategy is introduced to enhance label assignment by adaptively selecting high-quality samples across scales, improving boundary alignment and robustness to glottal variations.
    
    \item We design an adaptive feature disentanglement module based on deformable convolutions to decouple task-specific spatial features, enhancing geometric adaptability under occlusions and perspective shifts.
    
    \item The notable benefit of our approach in accuracy is verified on datasets and practical application scenarios, while preserving minimal model complexity (5MB) and superior inference speed ($>$62 FPS). 
\end{enumerate}

The remainder of this paper is organized as follows. We review related work in Section~\ref{related_works}. Section~\ref{method} describes the details of the proposed framework. The experimental datasets, details, and results are presented in Section~\ref{experiments}. Finally, Section~\ref{Discussion and conclusion} discusses and concludes the work and plans for the future.

\section{Related Works}
\label{related_works}
\subsection{Glottis Detection}
Glottis detection serves as a fundamental component in vision-assisted intubation, computer-aided airway diagnosis, and intelligent respiratory monitoring systems. Its accuracy directly influences the success rate of critical procedures such as endotracheal intubation and laryngeal pathology screening. 
Early investigations primarily adopted traditional image processing techniques, relying on handcrafted features and rule-based segmentation~\cite{gloger2014fully}. Andrade-Miranda et al.~\cite{andrade2015automatic} combined active contouring with watershed transformation to delineate the glottis, while Moccia et al.~\cite{moccia2017confident} proposed a texture-based classification model with an integrated confidence assessment. These methods advanced initial research but remained vulnerable to illumination changes, anatomical variability, and noise, limiting their reliability in real-world scenarios. 
With the development of deep learning, Convolutional Neural Networks (CNNs) quickly became the mainstream paradigm due to their ability to automatically learn high-level semantic features. Publicly available clinical datasets have played an important role in advancing glottis detection research. For example, the BAGLS benchmark~\cite{gomez2020bagls} provides a large-scale resource for glottis segmentation, while the dataset introduced by Laves et al.~\cite{Dataset1} offers additional laryngeal endoscopic images for evaluating CNN-based semantic segmentation. These datasets have collectively facilitated the training of sophisticated architectures. More recently, Transformer-based networks have been explored in medical imaging, aiming to capture global dependencies that CNNs inherently struggle to model. In parallel, lightweight detectors such as YOLACT~\cite{YOLACT} have also been applied to laryngeal and airway imaging, achieving real-time inference on clinical videos. Together, these developments have substantially advanced the field, yet the pursuit of solutions that can simultaneously ensure robustness, precision, and efficiency remains an ongoing challenge.

\subsection{General Object Detection}
Beyond medical applications, the broader field of object detection has seen rapid methodological progress, which provides a critical foundation for glottis detection research. Two-stage detectors such as Faster R-CNN~\cite{Faster-RCNN} integrate region proposal networks with detection heads, achieving high localization accuracy but at the expense of heavy computational overhead, which limits real-time deployment. In contrast, one-stage detectors including RetinaNet~\cite{retinanet}, CenterNet~\cite{centernet}, and RTMDet~\cite{lyu2022rtmdet} eliminate proposal generation to improve inference speed, but often exhibit reduced precision in boundary delineation and small-object localization. 
CNN-based models, while widely adopted, inherently rely on localized receptive fields and typically require deep hierarchical structures to capture global context, substantially increasing computational cost. More recently, Transformer-based approaches such as DETR~\cite{DETR} and its variants~\cite{Conditional_DETR,DAB_DETR,Deformable_DETR} introduced global attention mechanisms that alleviate CNN locality limitations and enable set-based prediction. Nevertheless, these models exhibit a strong dependence on large-scale pretraining corpora and suffer from quadratic attention complexity, amplifying computational demands and constraining their practicality in time-sensitive scenarios. Parallel efforts including GFL~\cite{GFL}, TOOD~\cite{TOOD}, and ATSS~\cite{ATSS} improved label assignment strategies to boost detection accuracy, but still fall short of reconciling the trade-off between efficiency and precision. 
Consequently, there is a pressing need for detection frameworks that are both lightweight and accurate, enabling reliable coordinate estimation under real-time constraints and thereby facilitating their integration into vision-guided intubation and other clinical procedures.

\subsection{Robotic Systems for Airway Management and Glottis Detection}
Driven by interdisciplinary integration, the combination of artificial intelligence (AI)-based visual recognition and robotic technology has been increasingly applied to airway management, emerging as a key frontier in this field. 
As early as 2012, Hemmerling et al.~\cite{hemmerling2012first} proposed the Kepler Intubation System, which consisted of a remote control center, a robotic arm, and a standard video laryngoscope. This pioneering system enabled intubation experiments under direct, indirect, and semi-automatic modes, marking the entry of robotic intubation into clinical exploration. To further improve glottic exposure during complex procedures, Rodney et al.~\cite{rodney2016robotic} developed the Modular Oral Retractor (MOR) in 2016, which enhanced operability by providing specialized auxiliary access. 
Building on these foundations, Cheng et al.~\cite{cheng2018intubot} presented the IntuBot system, which applies computer vision to identify airway landmarks such as the epiglottis, vocal folds and glottis, and links these outputs to robotic control for endotracheal tube guidance. Subsequently, Biro et al.~\cite{biro2020automated} proposed the Robotic Endoscope Automated Via Laryngeal Imaging for Tracheal Intubation system in 2020, designed to automatically drive the endoscope tip to the geometric center of the glottis once detected, thereby assisting clinicians with higher efficiency.
In parallel, existing surgical robotic platforms have also been extended to laryngeal applications. Systems such as the da Vinci Surgical System and the Flex Robotic System have been widely adopted in Transoral Robotic Surgery~\cite{rosello2020transoral}, particularly for vocal cord tumor resections. Although not originally designed for glottis detection, they provide high-definition three-dimensional visualization and dexterous instrument manipulation, greatly enhancing the surgeon’s ability to localize and operate around the glottis. Most recently, Zhang et al.~\cite{zhang2024lightweight} proposed a modular flexible robotic laryngoscope characterized by a lightweight and detachable design, integrated with an intelligent visual guidance module. Equipped with compact deep learning algorithms, this system achieves real-time glottis recognition and motion path planning, combining accuracy with portability, and is especially suitable for emergency and bedside applications.
While these robotic systems demonstrate significant advances in airway management and glottis detection capabilities, the complex deep learning models employed by current systems often demand substantial computational resources and memory, making real-time processing challenging on mobile or bedside devices.

\section{Algorithm}
\label{method}
In this section, we first introduce the overall real-time glottis detection framework. It employs a computationally efficient backbone and a hierarchical selection mechanism to alleviate gradient conflicts from fixed matching strategies in traditional detectors, enabling adaptive sample assignment through multi-scale feature fusion and dynamic threshold computation.

\begin{figure*}[!t]
\centering
\includegraphics[width=\textwidth]{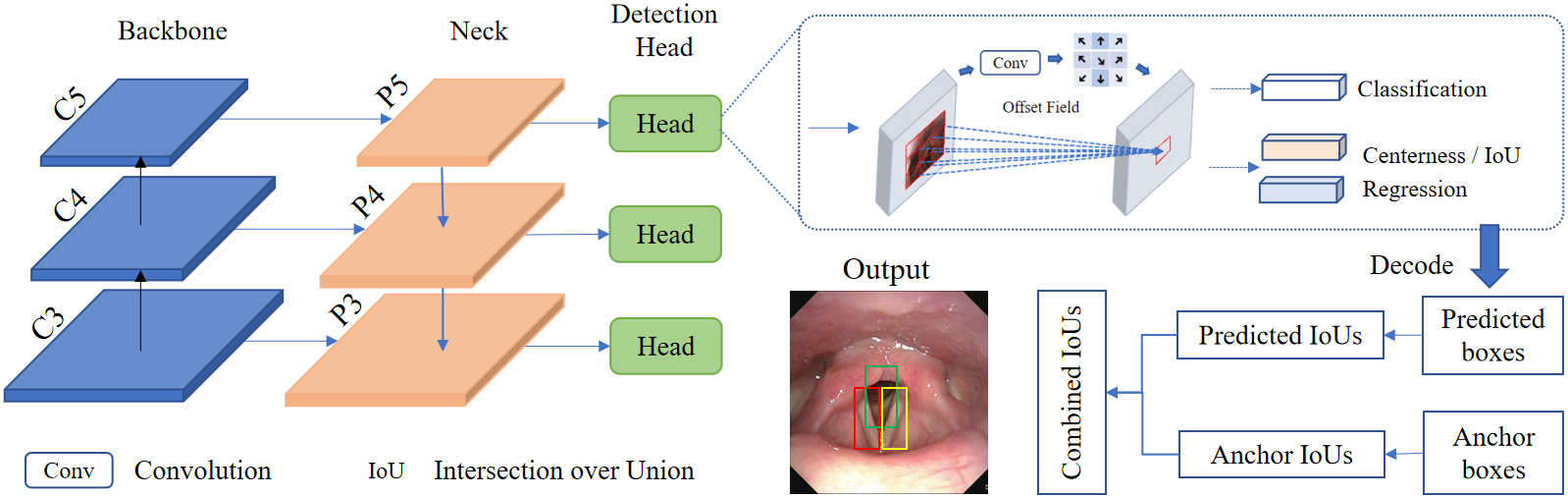}
\caption{Overview of the proposed Mobile GlottisNet. The input image is first processed by the backbone network to generate multi-scale feature maps, which are then fused through the FPN to produce enhanced multi-scale feature representations. Subsequently, the classification and regression tasks are decoupled, and dynamic label assignment strategies are applied to adjust anchor point allocation. The mean IoU of candidate boxes is calculated to set an adaptive threshold. Finally, convolutional kernel sampling positions are adjusted based on the Offset Field in the adaptive feature decoupling module, and the final detection results are derived from class scores and bounding box offsets.}
\label{overview}
\end{figure*}

\subsection{Overall Framework}

We present Mobile GlottisNet, a lightweight framework tailored for embedded glottis detection. As shown in Fig.~\ref{overview}, the network is optimized for both architectural efficiency and hardware adaptability, enabling accurate real-time inference on resource-constrained platforms.
To reduce computation, a lightweight backbone is adopted. A hierarchical selection mechanism is introduced to alleviate gradient conflicts in sample assignment, leveraging multi-level features and dynamic thresholds for adaptive classification and regression.
The fused features are processed by an adaptive feature decoupling module, which employs deformable convolutions to focus on key glottic regions and suppress visual occlusions. Final predictions are generated through a lightweight detection head.
This design achieves a balance between detection accuracy and computational efficiency, meeting the demands of low-power clinical deployment.

\subsection{Lightweight Backbone}
In nasal intubation scenarios, precise recognition and localization of anatomical structures are crucial. However, the limited computational capacity of medical devices often impedes the deployment of large models. To this end, we adopt MobileNetV3~\cite{MobileNetV3} as our lightweight backbone, which combines neural architecture search with hand-crafted design to balance performance and efficiency. 

We employ depthwise separable convolutions and inverted residual blocks, drastically reducing the number of parameters and floating-point operations (FLOPs) while maintaining competitive accuracy. To further enhance feature expressiveness, we integrate a lightweight Squeeze-and-Excitation (SE) module that recalibrates channel-wise responses by learning global context through average pooling.
Additionally, we utilize the hardware-friendly \textit{h-swish} activation function, which approximates the Swish function via piecewise linearity to reduce computational burden while preserving nonlinearity:
\begin{equation}
\text{h-swish}(x) = x \cdot \frac{1}{6}\text{ReLU6}(x + 3).
\end{equation}
This activation maintains strong performance while being better suited for edge devices, making it ideal for our deployment scenario.

\subsection{Dynamic Thresholding and Hierarchical Sample Allocation}
To enhance the feature representation in glottis detection tasks, we optimized the label allocation strategy for classification and regression tasks by combining candidate screening, dynamic thresholding, and hierarchical selection. By constructing topological mapping relationships in a high-dimensional feature space, accurately match candidate regions with true boundaries of anatomical structures. Accurately identifying the geometric relationship between the epiglottis, glottis, and tracheal entrance is crucial during nasal endotracheal intubation.The specific procedure is outlined in \textbf{Algorithm 1}

\begin{algorithm*}[!t]
\caption{Glottis Candidate Selection with Dynamic Thresholding}
\begin{tabular*}{\textwidth}{@{\extracolsep{\fill}}l@{}}
\textbf{Require}: 
- Predicted bounding boxes \( \{b_i\} \) across \( L \) FPN scales, \\
- Ground-truth class \( C_j \) for the \( j \)-th glottis, \\
- Hyperparameters: balance factor \( \lambda \), top-\( K \) count. \\
\textbf{Ensure}: Selected candidate set \( T_j \) for training classification/regression. \\
\hline
\textbf{1. Cost Matrix Calculation} \\
For each predicted box \( b_i \) and ground-truth \( j \): \\
\quad \( C_{ij} = -\lambda \cdot \log\left( \frac{e^{P_i^{C_j}}}{\sum_{k=1}^K e^{P_i^k}} \right) + (1 - \lambda) \cdot \left| 1 - \text{IoU}(b_i, b_j) \right| \) \\
where \( P_i^{C_j} \) is the predicted probability of \( b_i \) belonging to class \( C_j \). \\
\textbf{2. Multi-Scale Top-\( K \) Candidate Selection} \\
For each FPN level \( l \in \{1, 2, \dots, L\} \): \\
\quad Select top-\( K \) boxes with the smallest \( C_{ij} \): \\
\quad \( T_j^l = \arg\ \text{top-}K(C_{ij}) \) (for level \( l \)) \\
Combine candidates from all levels: \\
\quad \( T_j = \bigcup_{l=1}^L T_j^l \) \\
\textbf{3. Dynamic Threshold Filtering} \\
Compute threshold \( \tau \) using batch-level statistics of \( \{C_{ij}\} \). \\
Retain only candidates with \( C_{ij} < \tau \): \\
\quad \( T_j = \left\{ b_i \in T_j \mid C_{ij} < \tau \right\} \) \\
\textbf{Output}: Candidate set \( T_j \) for subsequent tasks. \\
\end{tabular*}
\end{algorithm*}

We establish a hierarchical dynamic threshold mechanism, and the algorithm only retains the candidate point set within the true glottal anatomical boundary, achieving sparse representation of the feature embedding space. A cost matrix is defined to quantify the matching quality between each predicted box and ground truth, combining classification probabilities and regression accuracy:
\begin{equation}
C_{ij} = - \lambda \cdot \log \left( \frac{e^{P_{i}^{C_{j}}}}{\sum_{k=1}^{K} e^{p_{i}^{k}}} \right) + (1 - \lambda) \cdot \left[ 1 - \text{IoU}(b_{i}, b_{j}) \right],
\end{equation}
where $P_{i}^{C_{j}}$ represents the predicted probability for the true category $C_j$. The algorithm selects $K$ prediction boxes that best align with the glottal structure across the multi-scale FPN, ensuring that all scale features participate in the learning of glottal contours:

\begin{equation}
T_{j} = \bigcup_{l = 1}^{L} \text{arg top-K}(C_{ij}).
\end{equation}

At the same time, by dynamically calculating the threshold through batch statistics, only candidates with matching quality exceeding the threshold are retained as positive samples to participate in subsequent classification and regression tasks. This strategy can accurately locate the glottal boundary in complex perspective changes under laryngoscopy, providing precise geometric guidance for intubation navigation and significantly reducing the risk of accidentally entering the esophagus during intubation.

\subsection{Adaptive Feature Disentanglement}
During the intubation process, variations in head posture and neck angle lead to substantial morphological changes in the glottis and its surroundings. To address this, we propose an \textit{adaptive feature disentanglement module} driven by deformable convolutions, enabling dynamic reconstruction of the feature space and alignment with key anatomical regions.

\begin{figure}[!t]
    \centering
    \includegraphics[width=3in]{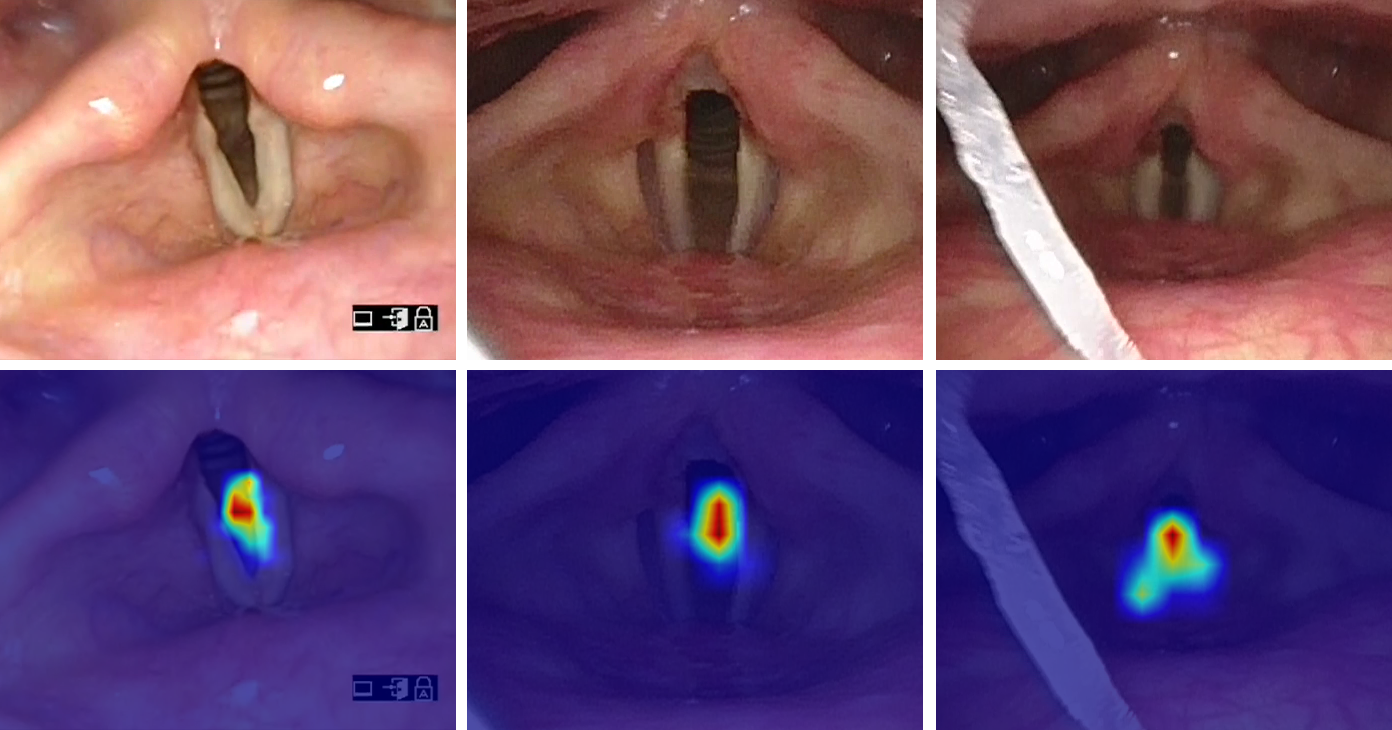}
    \caption{Attention heatmaps throughout the network. Feature disentanglement progressively refines spatial attention, culminating in accurate focus on the anatomical glottic aperture.}
    \label{feature}
\end{figure}

Given an input feature map $F \in \mathbb{R}^{H \times W \times C}$, a standard convolution samples values on a regular grid $\mathcal{R}$, while a deformable convolution dynamically adjusts sampling locations via a set of learned offsets $\Delta p \in \mathbb{R}^{k \times k \times 2}$. These offsets are generated by a lightweight sub-network $\Phi(F;\theta)$, predicting a 2D offset $\Delta p_q = (\Delta x_q, \Delta y_q)$ for each location $q \in \mathcal{R}$. The deformable convolution can then be formulated as:
\begin{equation}
F_{\text{cls/reg}}(p) = \sum_{q \in \mathcal{R}} W_{\text{cls/reg}}(q) \cdot F\left(p + q + \Delta p^{\text{cls/reg}}_q\right),
\end{equation}
where $W(q)$ is the kernel weight and $p$ denotes the output spatial location. The module predicts task-specific offsets independently for classification and regression. This facilitates task-aware spatial focus, enhancing the flexibility of feature representation and enabling spatial disentanglement. Furthermore, end-to-end trainability is ensured by bilinear interpolation, allowing gradient flow back to $\Phi$:
\begin{equation}
\frac{\partial F_{\text{disentangled}}}{\partial \Delta p_q} = W_q \cdot \frac{\partial F}{\partial(p + q + \Delta p_q)}.
\end{equation}

Consequently, the classification branch emphasizes global semantic structures (e.g., glottal shape), while the regression branch attends to boundary-level details (e.g., glottic edge localization). By integrating multi-scale feature pyramids and cross-layer supervision with dynamic weighting, the module achieves refined spatial alignment and semantic-detail fusion. The specific implementation process is depicted in \textbf{Algorithm 2}.

\begin{algorithm*}[ht]
\caption{Deformable Convolution Forward Computation Pipeline}
% 使用p列实现自动换行，宽度设为文本宽度（减去左右边距补偿）
\begin{tabular}{@{}p{\dimexpr\textwidth-2\tabcolsep}@{}}
\textbf{Input}: Feature map \( F \in \mathbb{R}^{H \times W \times C} \), regular grid \( \mathcal{R} \), convolutional kernel weights \( W \in \mathbb{R}^{k \times k \times C} \); \\
Parameters: Offset generation network parameters \( \theta \), uncertainty ratio \( \eta \); \\
1. Offset Generation: \\
\quad Compute offsets \( \Delta p^{\text{cls/reg}} = \Phi(F; \theta) \), where for all \( q \in \mathcal{R} \), 
\quad\( \Delta p_q^{\text{cls/reg}} = (\Delta x_q^{\text{cls/reg}}, \Delta y_q^{\text{cls/reg}}) \); \\
2. Deformable Sampling: \\
\quad For each output position \( p \), \( F_{\text{cls}}(p) = \sum_{q \in \mathcal{R}} W_{\text{cls}}(q) \cdot F\bigl(p + q + \Delta p_q^{\text{cls}}\bigr) \), and 
\quad\( F_{\text{reg}}(p) = \sum_{q \in \mathcal{R}} W_{\text{reg}}(q) \cdot F\bigl(p + q + \Delta p_q^{\text{reg}}\bigr) \); \\
3. Gradient Backpropagation Mechanism: \\
\quad The gradient satisfies \( \frac{\partial F}{\partial \Delta p_q} = W_q \cdot \frac{\partial F}{\partial \bigl(p + q + \Delta p_q\bigr)} \), ensuring end-to-end trainability 
\quad via bilinear interpolation; \\
4. Multi-scale Feature Fusion: \\
\quad \( F_{\text{final}} = \text{DynamicWeighting}\bigl(\text{FeaturePyramid}(F), F_{\text{cls}}, F_{\text{reg}}\bigr) \); \\
\textbf{Output}: Classification feature \( F_{\text{cls}} \), regression feature \( F_{\text{reg}} \). \\
\end{tabular}
\end{algorithm*}

To visualize the effectiveness, we provide attention maps in Fig.~\ref{feature}. The results demonstrate that our method localizes the glottic gap precisely while suppressing irrelevant contextual noise, yielding uniform and interpretable attention responses over the glottic aperture.

\section{Experiments}
\label{experiments}
In this section, we first describe the hardware system for robotic NTI and provide implementation details. Next, we compare our proposed Mobile GlottisNet  with the state-of-the-art (SOTA) methods. Finally, we present an ablation study and a discussion on experimental results.

\subsection{Hardware System}

\begin{figure}[!t]
\centering
\includegraphics[width=3in]{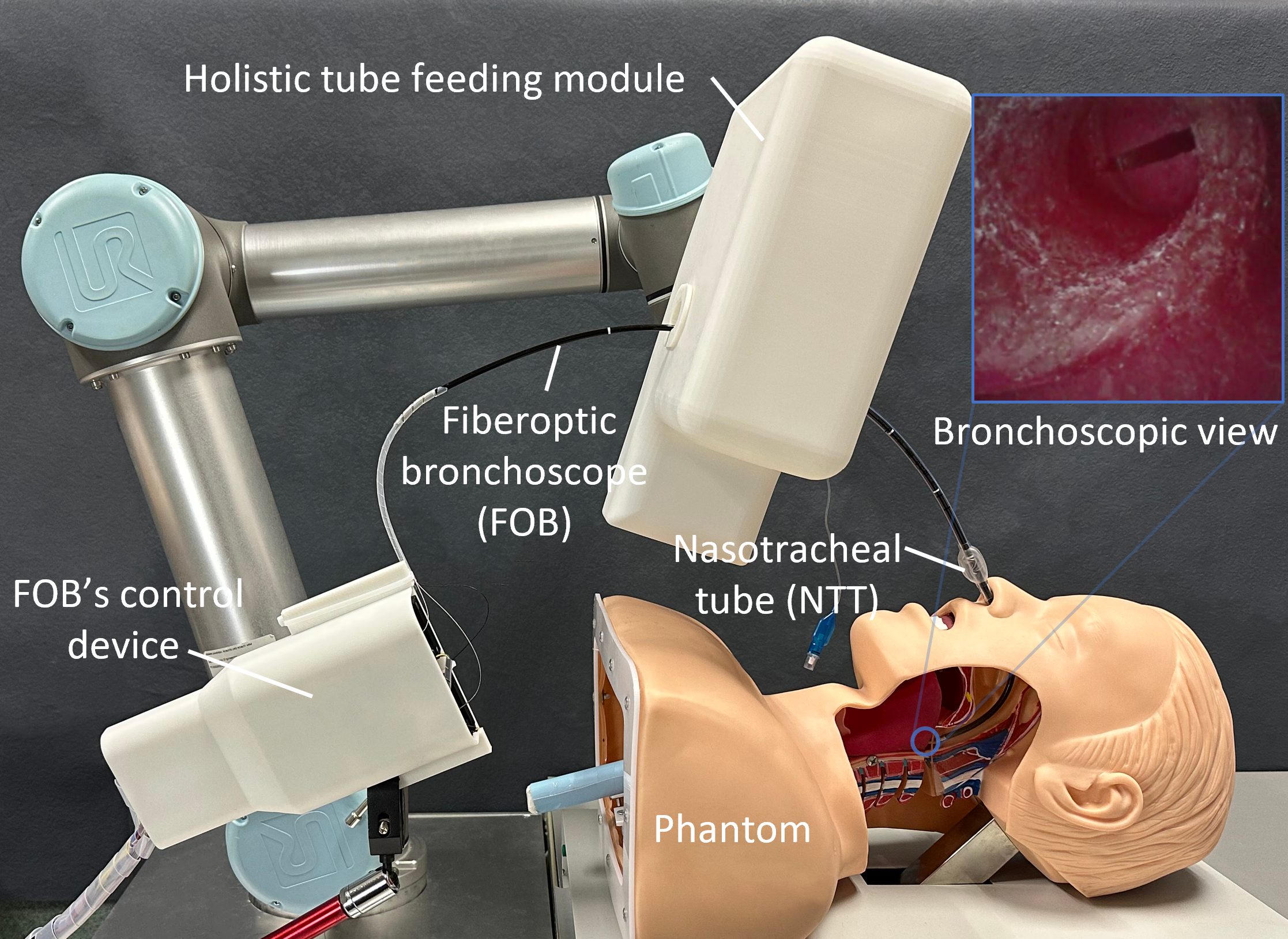}
\caption{The nasotracheal intubation robot system is an advanced medical system that improves intubation accuracy and minimizes human error. It utilizes a fiberoptic bronchoscope with a bendable robotic arm, dynamically adjusting insertion paths through real-time feedback control.}
\label{hardware}
\end{figure}

To validate the performance of the proposed framework in clinical scenarios, we performed experimental validation on the robotic NTI illustrated in Fig.~\ref{hardware}. This system was engineered to achieve automated insertion of a nasotracheal tube (NTT) through the nasal cavity, establishing a secure airway-oxygenator interface. The core control module enables precise articulation of the fiberoptic bronchoscope (FOB) distal tip with 360° servo-controlled maneuverability, while the integrated feed mechanism coordinates the translational movements of both FOB and NTT external components~\cite{hao2025variable}.

The robotic arm achieves precise teleoperation of the integrated feed mechanism and fiber-optic bronchoscope (FOB) through the following navigation control process. During nasal access, the FOB tip is guided by real-time bronchoscopic imaging to navigate from external phantom through nasal vestibule while maintaining a safe distance from the inferior turbinate mucosa to prevent epistaxis. The procedure subsequently advances through pharyngeal transit, during which adaptive path planning algorithms dynamically modify the insertion trajectory throughout the nasopharynx and oropharynx to accommodate anatomical variations in adult airways. Finally, glottic alignment is accomplished as the vision system automatically detects key anatomical landmarks, including anterior commissure, posterior glottic plate, and interarytenoid notch within simulated airway models, ensuring optimal positioning and procedural accuracy.

\subsection{Experimental Setup}
\subsubsection{Implementation Details}
We trained our proposed Mobile GlottisNet model using the SGD optimizer with a momentum of 0.9 and a weight decay of 0.05 to stabilize training and prevent overfitting. A cosine annealing learning rate schedule was adopted, starting from 0.01 and gradually decaying to zero. All ablation studies were conducted with a batch size of 128 over 36 epochs, and each experiment was repeated three times, with mean results reported to reduce the influence of randomness. 
To handle inter-subject glottic morphological variability, raw laryngoscopic images were uniformly resampled to $400 \times 400$ pixels using bicubic interpolation. All experiments were conducted on a workstation equipped with an Intel Core i9-10980XE CPU, 128\,GB of system memory, and a single NVIDIA RTX 3090 GPU. Inference latency was benchmarked on an NVIDIA Jetson Orin to assess edge-device deployment feasibility. The computational environment was implemented with MMDetection, using CUDA 11.6 and PyTorch 1.12 to ensure reproducibility.
For fair comparison, each baseline model was trained specifically for the glottis detection task, using either the original hyperparameter settings reported in its publication or our configuration when unspecified. All models shared the same input resolution, data preprocessing, and augmentation pipeline. Random seeds were fixed across runs, and model selection was based on the best validation accuracy achieved during training.

\subsubsection{Datasets}
To systematically evaluate the effectiveness and robustness of our method, we conducted comprehensive experiments on three complementary datasets. These datasets vary in scale, source, and acquisition settings, thereby ensuring applicability across diverse scenarios.

\begin{figure}[!t]
\centering
\includegraphics[width=3in]{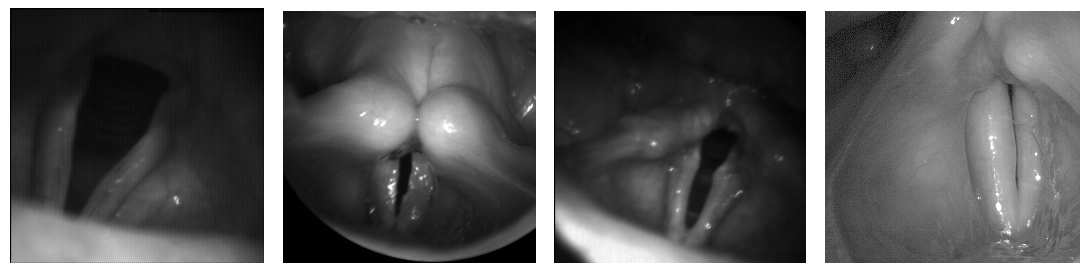}
\caption{Sample images from the Glottis dataset. The dataset includes high-speed videoendoscopic recordings from both healthy individuals and patients with laryngeal disorders, collected across different clinical settings and imaging devices. }
\label{Bagls}
\end{figure}

\textbf{Glottis}~\cite{gomez2020bagls}: The dataset comprises high-speed videoendoscopic recordings from 640 participants, including both healthy subjects and patients with laryngeal disorders. Recordings were acquired by multiple clinicians using heterogeneous endoscopic systems, resulting in pronounced variability in resolution, illumination, and anatomical presentation. Representative examples illustrating this variability are shown in Fig.~\ref{Bagls}. For compatibility with object detection, we converted the original pixel-level segmentation masks into bounding-box annotations and reformatted the labels to the COCO~\cite{lin2014coco} standard. After preprocessing, the dataset contains 55{,}750 training images and 3{,}500 testing images. Owing to its scale and heterogeneity, Glottis constitutes a strong benchmark for assessing the generalization ability of detection models under diverse imaging conditions.

\begin{figure}[!t]
    \centering
    \begin{subfigure}{0.8\linewidth}
        \centering
        \includegraphics[width=\linewidth]{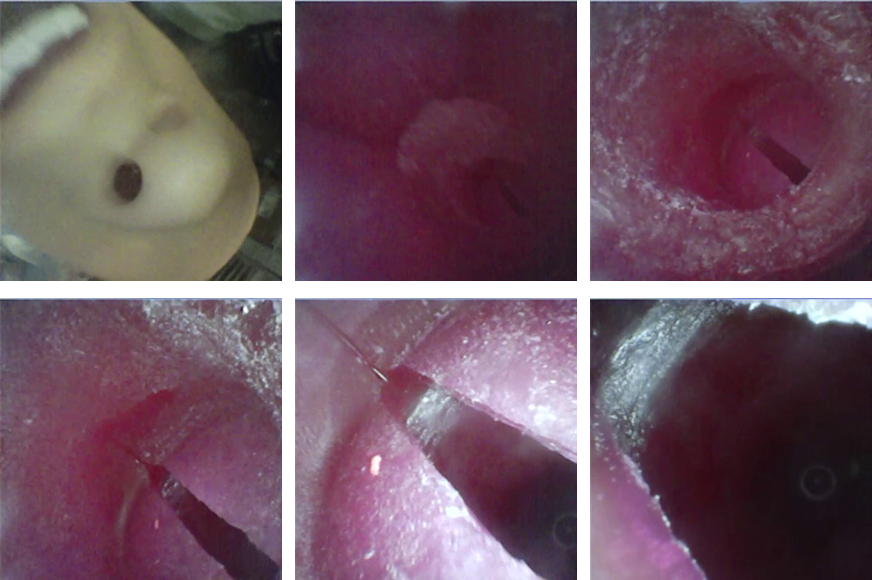}
        \caption{Sample images from the PID dataset}
        \label{fig:pid_samples}
    \end{subfigure}
    
    \begin{subfigure}{0.8\linewidth}
        \centering
        \includegraphics[width=\linewidth]{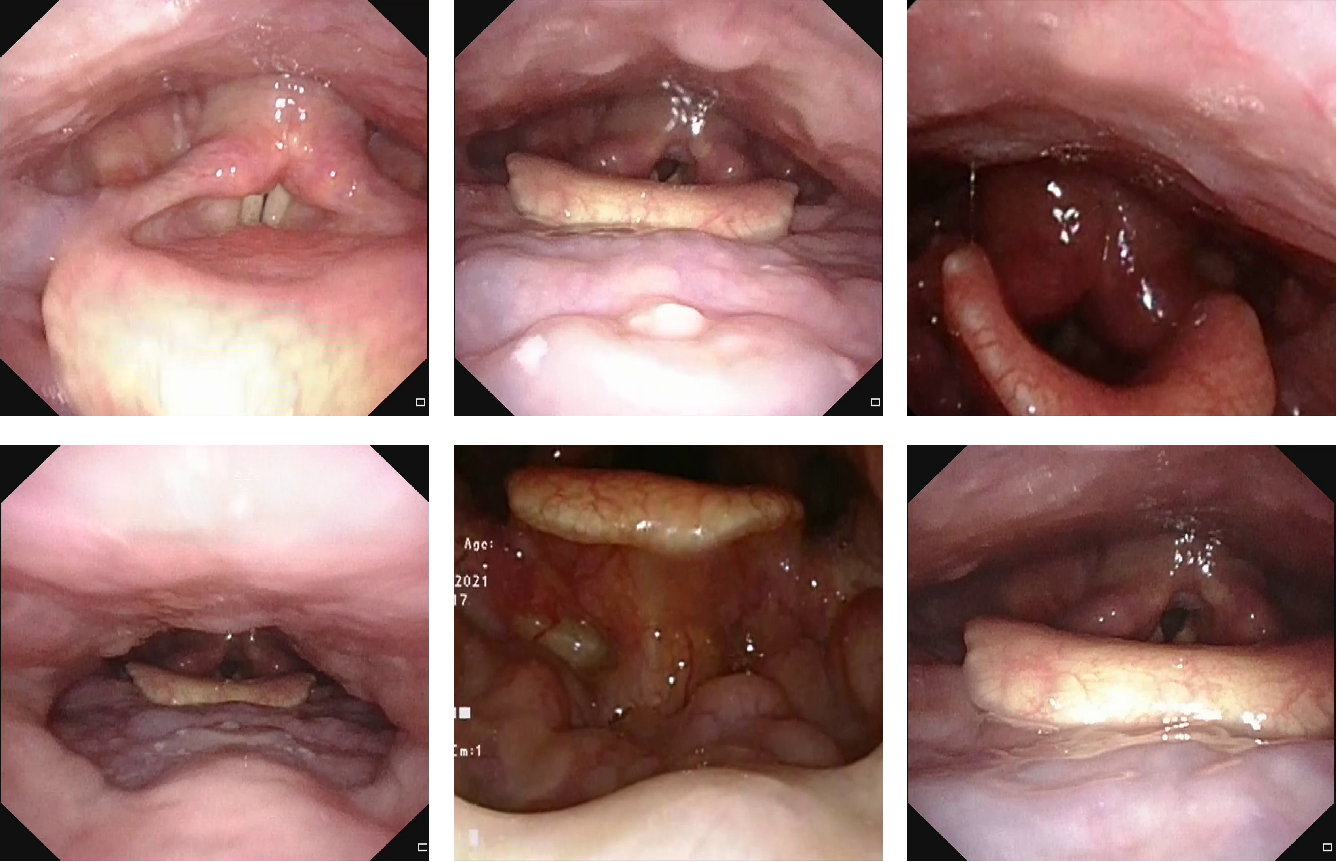}
        \caption{Sample images from the Clinical dataset}
        \label{fig:clinical_samples}
    \end{subfigure}
    \caption{During NTI, the glottis undergoes substantial scale variation under complex and low-light conditions, and the surrounding anatomy is intricate. These factors often degrade the accuracy of conventional detectors.}
    \label{fig:nti_datasets}
\end{figure}

\textbf{PID}~\cite{hao2024uaal}: To alleviate ethical constraints during early development, we collected the Procedure-Inspired Design (PID) dataset in a laboratory setting using our robotic NTI platform, which integrates a fiberoptic bronchoscope with real-time navigation and feedback control. The dataset contains 2{,}267 training images and 479 test images. We adopt a hybrid annotation scheme: bounding boxes are assigned to general airway structures, namely the nasal passage, glottis, and trachea, for coarse localization, whereas instance masks are provided for anatomically complex regions, including the nostrils, the glottic slit, and the glottic valves, to capture precise boundaries. Representative frames with both annotation types are shown in Fig.~\ref{fig:nti_datasets}(\subref{fig:pid_samples}). This combination enables models to learn spatial localization together with fine structural detail, facilitating robust training without patient involvement.

\textbf{Clinical}~\cite{hao2024uaal}: To assess performance under clinical conditions, we curated the Clinical dataset from flexible nasopharyngoscopy recordings at Singapore General Hospital. All procedures were performed by board-certified otolaryngologists using commercial flexible nasopharyngoscopes to ensure clinical realism and protocol consistency. The dataset includes 2{,}683 training images with 7{,}030 annotated structures and 1{,}131 validation images with 3{,}295 annotations. Annotations cover key anatomical landmarks, including the nostrils, epiglottis, glottic valves, and trachea, and the sequences capture the progression as the endoscope traverses the glottis into the lower airway. Representative frames with annotations are shown in Fig.~\ref{fig:nti_datasets}(\subref{fig:clinical_samples}). This dataset reflects clinical complexity and provides a rigorous benchmark for evaluating model robustness under realistic operating conditions.

\subsubsection{Evaluation Metrics}
To evaluate object detection performance, we adopt standard COCO metrics~\cite{coco}, including the mean Average Precision (mAP), as well as $AP_{50}$ and $AP_{75}$, which correspond to IoU thresholds of 0.5 and 0.75, respectively. A higher Intersection over Union (IoU) threshold indicates stricter requirements for localization accuracy.
In addition to accuracy, computational efficiency is assessed by inference speed, reported in frames per second (FPS). To better reflect real-world deployment scenarios, we report FPS on both a desktop GPU (denoted as FPS\textsubscript{dev}) and an edge device (denoted as FPS\textsubscript{edge}). We also consider the model size, which affects memory consumption and plays a critical role in determining the feasibility of deployment on low-power platforms.

\begin{table*}[!h]
  \centering
  \renewcommand{\arraystretch}{1.25} % 调整行高
  \caption{Comparison of accuracy with the state-of-the-art methods on the PID and clinical datasets.}
  \label{tab:sota}
  \begin{threeparttable}
    \resizebox{\textwidth}{!}{%
      \begin{tabular}{lcccccccccc}
        \toprule
        \multirow{2}{*}{\makecell[c]{Methods}} & 
        \multirow{2}{*}{\makecell[c]{Backbone}} & 
        \multicolumn{3}{c}{PID (\%)}  & 
        \multicolumn{3}{c}{Clinical (\%)}  & 
        \multirow{2}{*}{\makecell[c]{Model\\ size (MB)}} &  
        \multicolumn{2}{c}{Infer (FPS)}  \\ 
        \cmidrule(lr){3-5} \cmidrule(lr){6-8} \cmidrule(lr){10-11}
        &  & mAP  & $AP_{50}$    & $AP_{75}$   & mAP & $AP_{50}$    & $AP_{75}$  & & Device & Edge \\        
        \midrule       
        ATSS~\cite{ATSS}        & ResNet50 & 30.3 & 59.5 & 28.4 & 17.8 & 39.5 & 13.6 & 251 & 21.6 & 12.1  \\ 
        CenterNet~\cite{centernet}   & ResNet18 & 25.5 & 54.9 & 24.0 & 12.6 & 25.9 & 10.9 & 114 & 50.0  & 25.3 \\ 
        CentripetalNet~\cite{CentripetalNet}  & Hourglass104 & 28.0 & 49.2 & 26.5 & 5.3 & 9.9 & 4.7 & 2416 & 11.2 & 5.9 \\ 
        Faster RCNN~\cite{Faster-RCNN} & ResNet50 & 29.1 & 55.1 & 27.6 & 32.4  & 61.0 & 29.3 & 323 & 51.3 & 26.1 \\ 
        GFL~\cite{GFL}         & ResNet101 & 33.1 & 61.3 & 32.7 & 34.9 & 61.5 & 35.7 & 401 & 30.1 & 16.4   \\
        RetinaNet~\cite{retinanet} & ResNet50 & 29.6 & 53.0 & 28.9 & 37.6 & 65.6 & 39.9 & 533 & 21.6 & 12.2 \\
        Reppoints~\cite{Reppoints} & ResNet50 & 21.7 & 48.5 & 16.7 & 15.5 & 33.5 & 12.2 & 288 & 38.8 & 19.7 \\
        RTMDet~\cite{lyu2022rtmdet} & CSPNeXt & 33.6 & 61.1 & 31.7 & 34.0 & 59.6 & 34.8 & 149 & 37.7 & 19.1  \\
        Sparse RCNN~\cite{SparseRCNN} & ResNet50 & 11.3 & 21.6 & 10.4 & 21.0 & 37.7 & 20.7 & 1244 & 29.6 & 16.2 \\
        TOOD~\cite{TOOD} & ResNet50 & 32.1 & 62.5 & 30.2 & 37.0 & 60.3 & 39.9 & 250 & 28.3 & 15.7 \\ 
        VarifocalNet~\cite{VfNet} & ResNet50 & 29.3 & 56.3 & 28.1 & 28.8 & 56.5 & 26.5 & 255 & 19.8 & 10.6 \\ 
        DETR~\cite{DETR}        & ResNet50 & 6.7  & 17.3 & 5.2  & 9.2 & 21.6 & 6.4 & 497 & 41.6 & 21.1 \\ 
        Conditional DETR~\cite{Conditional_DETR}   & ResNet50 & 15.6 & 37.3 & 10.2 & 10.5 & 34.1 & 2.6  & 512 & 34.4 & 17.9  \\ 
        DAB DETR~\cite{DAB_DETR}    & ResNet50 & 18.4 & 44.2 & 13.9 & 18.3 & 43.5 & 13.5 & 515 & 22.4  & 12.6 \\ 
        Deform. DETR~\cite{Deformable_DETR}& ResNet50 & 25.7 & 52.0 & 21.7 & 28.7 & 58.9 & 27.3 & 474 & 25.5 & 13.8 \\ 
        \midrule
        Mobile GlottisNet (Ours)       & MobileNetv3  & 33.3 & 57.5 & 35.1 & 35.5 & 66.8 & 33.9 & 33 & 53.5 & 26.6 \\  
        Mobile GlottisNet* (Ours)      & MobileNetv3  & 33.6 & 64.3 & 30.5 & 29.9 & 59.6 & 27.3 & 5  & 62.1 & 33.2 \\ 
        \bottomrule
      \end{tabular}
    }% end resizebox
    \begin{tablenotes}
      \footnotesize
      \item[*] The channel of FPN in neck can be further reduced to achieve a better trade-off.
      \item[] This adjustment may vary depending on the deployment scenario.
    \end{tablenotes}
  \end{threeparttable}
\end{table*}

\subsection{Comparison with State-of-the-Art Methods}
To comprehensively evaluate the performance of Mobile GlottisNet, we conducted comparative experiments on the PID and Clinical datasets, as shown in Table~\ref{tab:sota}, and further validated its generalization ability on the large-scale public Glottis dataset, as shown in  Table~\ref{SOAT}. The comparisons cover CNN-based detectors, transformer-based frameworks, and several efficient architectures. The results demonstrate that Mobile GlottisNet achieves the best balance between accuracy and efficiency, confirming its robustness and superiority across datasets.
\subsubsection{PID and Clinical}
As shown in Table~\ref{tab:sota}, different detection paradigms display clear differences in accuracy, localization under strict IoU thresholds, and computational efficiency. Anchor-based one-stage detectors such as GFL~\cite{GFL} and TOOD~\cite{TOOD} achieve high mAP on both datasets and maintain a relatively balanced relation between $AP_{50}$ and $AP_{75}$, which indicates stable localization at higher overlaps, although their inference speed and parameter budgets remain moderate. Anchor-free approaches such as CenterNet~\cite{centernet} obtain strong $AP_{50}$ yet drop markedly at $AP_{75}$, revealing limitations in boundary regression for small and morphologically variable glottic structures. DETR based transformer detectors~\cite{DETR,Deformable_DETR} underperform on this task, which appears to stem from one to one matching with sparse positive supervision, insufficient high resolution feature support for precise alignment, and score calibration dominated by the no object class, thereby limiting localization in endoscopic imagery.

Cross-domain comparisons further show that several anchor-based methods such as RetinaNet~\cite{retinanet} and TOOD~\cite{TOOD} remain relatively stable when transferring from the phantom domain to the clinical domain, whereas models relying on global attention or keypoint grouping degrade substantially. For example, CentripetalNet~\cite{CentripetalNet} exhibits a marked decline from PID to Clinical, indicating limited adaptability to complex real anatomical conditions. Two-stage detectors such as Faster R-CNN~\cite{Faster-RCNN} can reach competitive clinical mAP, yet their complex structures yield large parameter counts and higher latency, which constrains deployment on resource-constrained platforms; Sparse R-CNN~\cite{SparseRCNN} shows a similar trend. Among recent efficient designs, RTMDet~\cite{lyu2022rtmdet} improves the balance between accuracy and speed relative to earlier one-stage models such as ATSS~\cite{ATSS} and VarifocalNet~\cite{VfNet}, though performance at stricter IoU thresholds remains behind the best results.

Against this background, Mobile GlottisNet attains consistently strong performance at high IoU and sustains real-time throughput with a compact footprint. The MobileNetv3 variant offers competitive accuracy with a model size of about 33 MB and device-level throughput above 50 FPS, while the reduced-channel variant further compresses the model to 5 MB and maintains real-time speed on both device and edge platforms. Compared with efficiency-oriented baselines such as RTMDet~\cite{lyu2022rtmdet} and speed-oriented anchor-free designs such as CenterNet~\cite{centernet}, Mobile GlottisNet provides a more favorable joint outcome in accuracy and FPS, indicating a practical balance among precision, cross-dataset robustness, and deployability for clinical nasotracheal intubation.

\begin{table*}[!t]
  \centering
  \renewcommand{\arraystretch}{1.3}
  \caption{Comparison of accuracy with the state-of-the-art methods on the Glottis dataset.}
  \label{SOAT}
  \begin{threeparttable}
    % 调整列间距，避免表格过宽或过稀
    % \setlength{\tabcolsep}{6pt}
    % 缩放到版心宽度
    % \resizebox{\textwidth}{!}{%
      \begin{tabular}{l l ccc c c}
        \toprule
        \multirow{2}{*}{\makecell[c]{Methods}} &
        \multirow{2}{*}{\makecell[c]{Backbone}} &
        \multicolumn{3}{c}{Glottis (\%)} &
        \multirow{2}{*}{\makecell[c]{Model\\ size (MB)}} &
        \multirow{2}{*}{\makecell[c]{Inference\\(FPS)}} \\
        \cmidrule(lr){3-5}
        & & $mAP$ & $AP_{50}$ & $AP_{75}$ & & \\
        \midrule
        ATSS~\cite{ATSS} & ResNet50 & 53.7 & 84.8 & 60.1 & 251   & 21.6 \\
        BoxInst~\cite{BoxInst} & ResNet101 & 53.3 & 82.7 & 58.5 & 447 & 45.7 \\
        CondInst~\cite{CondInst} & ResNet50 & 41.1 & 79.1 & 36.9 & 279 & 46.8 \\
        CMask RCNN~\cite{Cascade_RCNN} & ResNet50 & 55.9 & 83.2 & 62.9 & 201   & 16.4 \\
        Conditional DETR~\cite{Conditional_DETR} & ResNet50 & 7.1  & 21.9 & 3.0  & 512   & 34.4 \\
        DAB DETR~\cite{DAB_DETR} & ResNet50 & 8.7  & 23.7 & 5.0  & 515   & 22.4 \\
        Deform. DETR~\cite{Deformable_DETR} & ResNet50 & 38.4 & 74.7 & 35.3 & 474   & 25.5 \\
        DETR~\cite{DETR} & ResNet50 & 4.6  & 15.2 & 1.7  & 497   & 41.6 \\
        GFL~\cite{GFL} & ResNet101 & 54.2 & 83.8 & 60.7 & 401   & 30.1 \\
        Mask2Former~\cite{Mask2Former} & ResNet50 & 39.8 & 72.3 & 39.0 & 288   & 38.8 \\
        Mask RCNN~\cite{maskrcnn} & ResNet50 & 51.2 & 79.1 & 57.1 & 129   & 31.5 \\
        Mask RCNN~\cite{maskrcnn} & Swin-T   & 53.8 & 79.4 & 62.4 & 572 & 38.2 \\
        MS RCNN~\cite{Mask-Scoring-RCNN} & ResNet50 & 48.2 & 75.9 & 53.9 & 198   & 17.5 \\
        PointRend~\cite{pointrend} & ResNet50 & 50.2 & 78.4 & 57.2 & 447 & 41.4 \\
        QueryInst~\cite{queryinst} & ResNet50 & 53.4 & 87.0 & 57.5 & 209 & 30.8 \\
        RTMDet\_s~\cite{lyu2022rtmdet} & CSPNeXt  & 55.4 & 87.5 & 63.6 & 149   & 37.7 \\
        YOLACT~\cite{YOLACT} & ResNet50 & 38.5 & 71.7 & 40.1 & 114   & 50.0 \\
        Ours & MobileNetv3 & \textbf{62.7} & \textbf{91.6} & \textbf{72.7} & \textbf{33}    & \textbf{53.5} \\
        \bottomrule
      \end{tabular}
    % }% end resizebox
  \end{threeparttable}
\end{table*}

\subsubsection{Glottis}
As shown in Table~\ref{SOAT}, two-stage detectors exemplified by Cascade Mask R-CNN~\cite{Cascade_RCNN} deliver competitive overall accuracy and strong high-IoU localization, yet their computational cost and latency are substantial, which limits suitability in time-critical settings. One-stage CNN detectors such as RTMDet\_s~\cite{lyu2022rtmdet}, ATSS~\cite{ATSS}, and GFL~\cite{GFL} offer a more balanced trade-off between accuracy and efficiency than two-stage counterparts and maintain strong $AP^{\text{box}}_{50}$, but performance still deteriorates under stricter IoU thresholds. Transformer-based detectors from the DETR family, including DETR~\cite{DETR}, Conditional DETR~\cite{Conditional_DETR}, DAB-DETR~\cite{DAB_DETR}, and Deformable DETR~\cite{Deformable_DETR} yield the weakest box-level results on this benchmark; even with deformable attention, convergence and matching on small glottic targets remain challenging under limited data. Meanwhile Mobile GlottisNet achieves strong accuracy across $mAP$, $AP^{\text{box}}{50}$, and $AP^{\text{box}}{75}$, with particularly pronounced gains at high IoU that evidence reliable fine-grained boundary localization for small and morphologically variable structures. At the same time, it sustains real-time throughput of 53.5 FPS with a compact footprint of approximately 33 MB. Compared with RTMDet\_s~\cite{lyu2022rtmdet}, it attains higher $mAP$ at faster speed with fewer parameters; compared with the strongest two-stage baseline Cascade Mask R-CNN~\cite{Cascade_RCNN}, it achieves higher $AP^{\text{box}}_{75}$ without incurring additional latency. Overall, Mobile GlottisNet offers a superior balance between detection accuracy and deployability.

\subsection{Ablation Studies}
\subsubsection{Backbone}
We conducted a systematic ablation study across multiple backbone architectures to analyze the trade-off between accuracy and efficiency in glottis detection. As shown in Table~\ref{tab:backbone_pid}, ResNet101 achieves the highest accuracy; however, its large model size and slow inference make it unsuitable for real time deployment on resource constrained platforms. ResNet50 and Swin-T offer competitive accuracy, yet their heavy parameterization and high latency lead to an unfavorable balance between accuracy and efficiency.

\begin{table}[!t]
  \renewcommand{\arraystretch}{1.15}
  \centering
  \caption{Performance of various backbones on the PID dataset.}
  \label{tab:backbone_pid}
  \setlength{\tabcolsep}{0.4mm}{
  \begin{tabular}{lccccc} % ← 6 列
    \toprule
    Backbone & mAP & $AP_{50}$ & $AP_{75}$ & \makecell[c]{Inference\\(FPS)} & \makecell[c]{Model\\ Size (MB)} \\
    \midrule
    ResNet101    & 38.8 & 72.5 & 35.4 & 26.6 & 205 \\
    ResNet50     & 33.3 & 66.3 & 30.6 & 31.5 & 129 \\
    ResNet18     & 30.1 & 61.6 & 29.2 & 40.2 & 77  \\
    ShuffleNet   & 18.7 & 42.4 & 11.6 & 65.4 & 73  \\
    Swin-T       & 35.3 & 63.7 & 28.6 & 21.2 & 285 \\
    EfficientNet & 28.7 & 54.4 & 21.9 & 51.4 & 92  \\
    MobileOne    & 27.1 & 51.5 & 20.8 & 62.5 & 41  \\
    MobileNetv3  & 33.3 & 57.5 & 35.1 & 53.5 & 33  \\
    MobileNetv3* & 33.6 & 64.3 & 30.5 & 62.1 & 5   \\
    \bottomrule
  \end{tabular}}
\end{table}

Lightweight backbones generally provided faster inference but showed varying levels of effectiveness. ShuffleNet and MobileOne achieved high frame rates but suffered significant drops in accuracy, revealing the limitations of overly aggressive lightweighting. ResNet18 performed better in accuracy but remained relatively large, reducing its appeal as a truly lightweight option.  
In contrast, MobileNetv3 struck the most effective balance, combining strong accuracy with compact size and real-time speed. Its optimized variant, MobileNetv3*, further compressed the model to only a few megabytes while maintaining stable accuracy and even higher efficiency. These results highlight MobileNetv3 as the most practical backbone for deployment in portable or embedded NTI systems, offering a robust balance between accuracy, efficiency, and adaptability under strict hardware constraints.

\subsubsection{Effect of TopK in Dynamic Thresholding}
Table~\ref{tab:topk} presents the influence of different TopK settings in dynamic thresholding and hierarchical sample allocation on the PID dataset. With very small TopK values such as 1 and 3, the detector fails to collect enough positive samples, leading to limited accuracy. As TopK increases, performance steadily improves, with clear gains observed when the value reaches 7, where the model achieves the best results. When TopK is further enlarged to 17, however, excessive candidate samples introduce noise and weaken supervision, resulting in performance degradation. These findings highlight the importance of selecting an appropriate TopK to balance sufficient positive sampling and noise suppression. Considering both accuracy and stability, TopK of 7 is adopted as the default setting in our subsequent experiments.

\begin{table}[!t]
  \renewcommand{\arraystretch}{1.1}
  \centering
  % \small
  \caption{Comparison of TopK in dynamic thresholding \& hierarchical sample allocation on PID dataset.}
    \label{tab:topk}
    \setlength{\tabcolsep}{0.9mm}{
    \begin{tabular}{cccc||cccc}
    \toprule
        TopK& 
        $mAP$& 
        $AP_{50}$& 
        $AP_{75}$&
        TopK& 
        $mAP$& 
        $AP_{50}$& 
        $AP_{75}$
        \\    \midrule
    1   & 27.4  & 51.0  & 26.7  & 11 & 33.5 & 63.0 & 32.5  \\
    3   & 32.3  & 57.0  & 31.5  & 13 & 30.6 & 58.6 & 30.0  \\
    5   & 31.7  & 57.5  & 31.8  & 15 & 30.2 & 57.2 & 27.9   \\
    7   & 35.0  & 63.4  & 34.6  & 17 & 32.0 & 59.6 & 31.1   \\
    9   & 33.3  & 66.3  & 30.6  & 19 & 35.7 & 66.0 & 35.5 \\ 
    \bottomrule
    \end{tabular}}
\end{table}

\subsubsection{Channel}
To systematically optimize the feature representation capability under computational constraints, we conducted ablation experiments on the number of channels in the neck. In Table~\ref{tab:fpn}, the model size gradually decreases as the number of channels decreases from 256 to 8. When the number of channels is 128, the accuracy reaches its highest value, which is 68.2 in $AP_{50}$. This configuration maximizes feature discriminability for subtle glottal structures while avoiding over-parameterization.

The 64 channels maintain competitive accuracy but further improve inference speed to 43.9 FPS and reduce model size to 97.3 MB, demonstrating graceful degradation. As the number of channels diminishes to 8, although the model size gradually decreases, the accuracy decreases significantly. This proves that when the channel is too small, glottal feature information will also be lost.
For high-precision phantom studies, 256 channels maximize $mAP$, prioritizing localization rigor. For latency-sensitive clinical deployments, 64 channels offer 43.9 FPS with minimal accuracy loss, meeting real-time intubation requirements. To ensure the tradeoff between accuracy and model size, we change the channel towards different datasets.
\begin{table}[!t]
  \renewcommand{\arraystretch}{1.1}
  \centering
  % \small
  \caption{Performance of Mobile GlotissNet with MobileNetv3 in various channels on PID dataset.}
    \label{tab:fpn}
    \setlength{\tabcolsep}{0.6mm}{
    \begin{tabular}{cccccc}
    \toprule
        \multirow{2}{*}{{Channels}}& 
        \multirow{2}{*}{{$mAP$}}& 
        \multirow{2}{*}{{$AP_{50}$}}& 
        \multirow{2}{*}{{$AP_{75}$}}&
        \multirow{2}{*}{\makecell[c]{Infer\\ (FPS)}}& 
        \multirow{2}{*}{\makecell[c]{Model Size \\(MB)}}
        \\
           & & & & & \\
    \midrule
    256   & 33.3  & 66.3  & 30.6  & 31.5 & 129.1  \\
    128   & 32.4  & 68.2  & 26.8  & 36.4 & 104.1 \\
    64    & 32.7  & 66.6  & 27.9  & 43.9 & 97.3  \\
    32    & 28.2  & 57.1  & 25.1  & 48.6 & 95.3  \\
    16    & 25.7  & 54.0  & 20.8  & 50.5 & 94.7  \\
    8     & 20.5  & 45.7  & 16.5  & 51.7 & 94.5  \\
    \bottomrule
    \end{tabular}}
\end{table}

\begin{figure}[!t]
  \centering
  % 紧凑些的列间距
  \setlength{\tabcolsep}{0.6pt}
  \renewcommand{\arraystretch}{1}

  % ---------- (a) Glottis ----------
  \begin{subfigure}{0.95\linewidth} % 子图整体略收窄
    \centering
    \begin{tabular}{
      @{} >{\centering\arraybackslash}m{0.12\linewidth}
          >{\centering\arraybackslash}m{\dimexpr\linewidth-0.12\linewidth-2\tabcolsep\relax} @{}} % 精确扣除列间隙
      {\scriptsize\bfseries \makecell[c]{Ground\\Truth}} & \includegraphics[width=\linewidth]{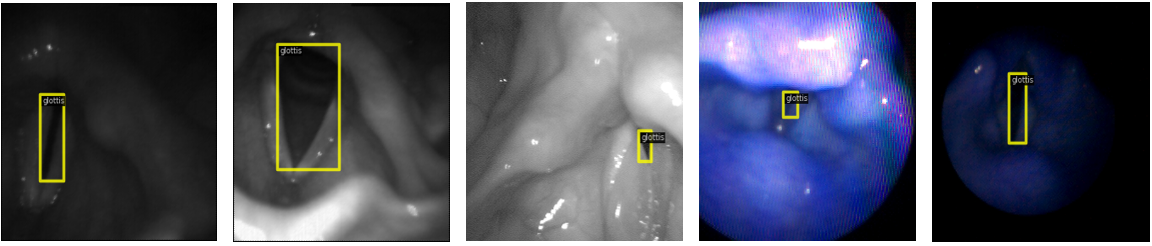} \\
      {\scriptsize\bfseries \makecell[c]{Prediction}}    & \includegraphics[width=\linewidth]{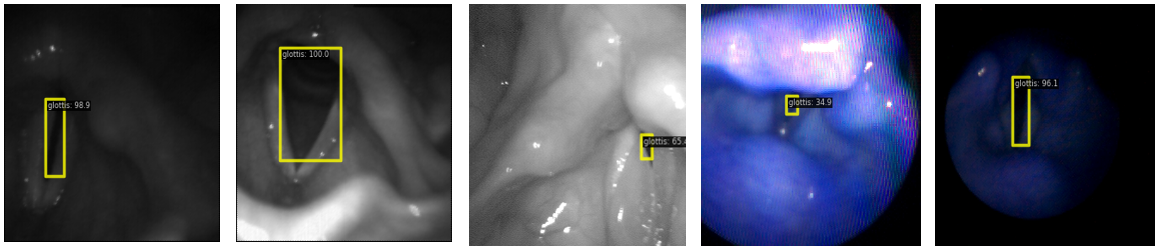} \\
    \end{tabular}
    \captionsetup{skip=0pt}
    \caption{\scriptsize Sample images from the glottis dataset}
    \label{fig:glottis_samples}
  \end{subfigure}

  \vspace{2pt}

  % ---------- (b) Real / Clinical ----------
  \begin{subfigure}{0.95\linewidth}
    \centering
    \begin{tabular}{
      @{} >{\centering\arraybackslash}m{0.12\linewidth}
          >{\centering\arraybackslash}m{\dimexpr\linewidth-0.12\linewidth-2\tabcolsep\relax} @{}} 
      {\scriptsize\bfseries \makecell[c]{Ground\\Truth}} & \includegraphics[width=\linewidth]{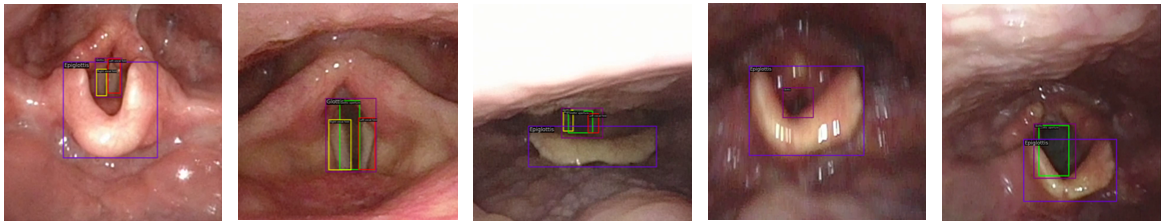} \\
      {\scriptsize\bfseries \makecell[c]{Prediction}}    & \includegraphics[width=\linewidth]{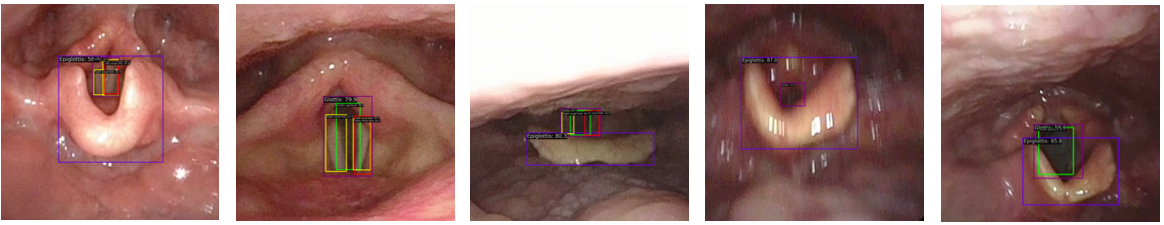} \\
    \end{tabular}
    \captionsetup{skip=0pt}
    \caption{\scriptsize Sample images from the clinical dataset}
    \label{fig:real_samples}
  \end{subfigure}

  \vspace{2pt}

  % ---------- (c) Phantom ----------
  \begin{subfigure}{0.95\linewidth}
    \centering
    \begin{tabular}{
      @{} >{\centering\arraybackslash}m{0.12\linewidth}
          >{\centering\arraybackslash}m{\dimexpr\linewidth-0.12\linewidth-2\tabcolsep\relax} @{}} 
      {\scriptsize\bfseries \makecell[c]{Ground\\Truth}}    & \includegraphics[width=\linewidth]{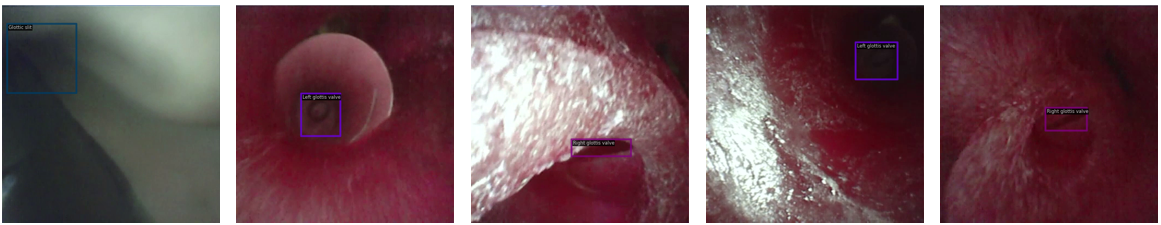} \\
      {\scriptsize\bfseries \makecell[c]{Prediction}}       & \includegraphics[width=\linewidth]{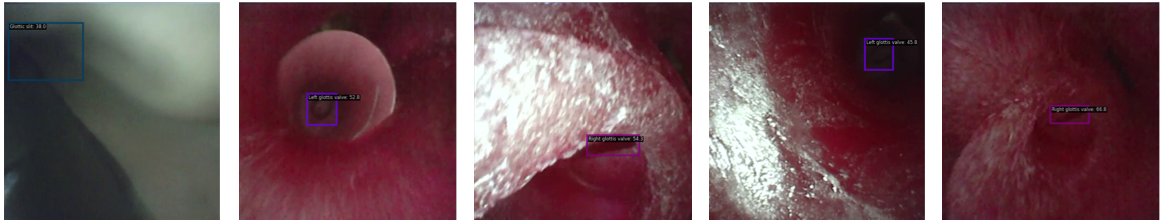} \\
      {\scriptsize\bfseries \makecell[c]{Video\\Detection}} & \includegraphics[width=\linewidth]{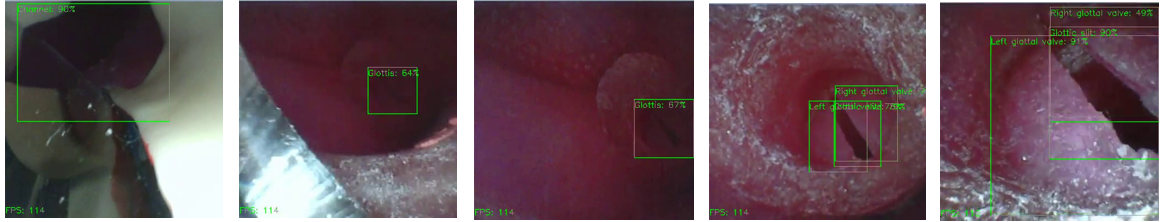} \\
    \end{tabular}
    \captionsetup{skip=0pt}
    \caption{\scriptsize Sample images from the phantom dataset}
    \label{fig:phantom_samples}
  \end{subfigure}

  \caption{\scriptsize Qualitative comparison among ground-truth labels, model predictions, and video detection results across glottis, clinical, and phantom datasets.}
  \label{fig:dataset_samples}
\end{figure}

\subsection{Qualitative results} 
As illustrated in Fig.~\ref{fig:dataset_samples}, our method achieves stable performance across glottis, clinical, and PID datasets, demonstrating strong robustness under adverse imaging conditions such as motion blur and anatomical variations. Even with complex backgrounds, the predicted bounding boxes remain well aligned with the ground truth, confirming the model’s ability to generalize beyond controlled laboratory settings.  
In particular, Fig.~\ref{fig:nti_datasets}(\subref{fig:phantom_samples}) highlights deployment on a robotic intubation platform. The visualizations are extracted from real-time video streams during phantom-based NTI procedures, where the model consistently maintained over 90 FPS on an embedded computing module. This validates that Mobile GlottisNet not only performs accurately in offline benchmarks but also operates in real-world robotic systems, enabling continuous glottis tracking and real-time decision support, thereby bridging the gap between algorithm development and clinical translation.

\section{Discussion and conclusion}
\label{Discussion and conclusion}
This paper introduced Mobile GlottisNet, a compact model for glottis detection in NTI that combines accurate localization, fast inference, and a small memory footprint. Across the PID, Clinical, and Glottis datasets, the model maintains reliable detection of small and morphologically variable glottic structures while still meeting device level and edge level timing budgets. Rather than trading accuracy for speed or compressing the network at the expense of stability, the design can yield a detector that is precise enough for clinical use and practical for time critical airway management.
Our hierarchical dynamic thresholding regularizes positive assignment under class imbalance and small-object conditions. When viewpoint changes and tissue motion induce non-rigid deformation, deformable convolution-based spatial alignment improves localization. The dynamic weighting of the cross-layers reconciles the semantic context with the fine structural detail on the scales. Taken together, these principles can inform detector design for laryngeal, airway, and other endoscopic applications, and they remain compatible with deployment on embedded and edge hardware where timing and memory budgets are tight. In robot-assisted visual intubation, this combination enables on-device perception that interfaces with tracking, view stabilization, and downstream control without reliance on external workstations.

Despite these advantages, several limitations remain. Severe occlusion by tissue or reflective fluids can disrupt box regression. The present study focuses on frame based inference and therefore does not exploit temporal continuity that is available in endoscopic video. In addition, generalization can drift across institutions and devices because optics, sensors, and protocols are not identical, and the system does not yet expose calibrated uncertainty that would assist safety aware decision making with human supervision. 
Future work will therefore incorporate occlusion aware decoding and targeted augmentation that reproduces specular highlights and secretions, extend the framework to video with temporal reasoning through joint detection and tracking and motion aware feature aggregation, and improve cross site robustness through semi supervised and self supervised pretraining, test time adaptation, and lightweight domain alignment. For deployment we will pursue hardware aware compression, including quantization and structured sparsity, so that latency and energy consumption continue to decrease while high threshold localization is preserved. We also plan to report confidence calibration and user facing reliability summaries, and to design prospective evaluations that measure usability and impact within clinical and robotic workflows.

\sloppy
\section*{Data availability}
Our experimental evaluation is conducted on three datasets: 
The BAGLS dataset is available at  \url{https://www.x-mol.com/paperRedirect/1274088604830298112}.
The PID and Clinical datasets will be made available upon publication at \url{https://doi.org/10.6084/m9.figshare.26342779.v3}.

\bibliography{ref}  % 此处 ref 即你的 .bib 文件名，无需加 .bib 后缀
\end{document}